\documentclass[10pt,twocolumn,letterpaper]{article}

\usepackage{cvpr}
\usepackage{times}
\usepackage{epsfig}
\usepackage{graphicx}
\usepackage{amsmath}
\usepackage{amssymb}

\usepackage{url}
\usepackage{booktabs}
\usepackage{multirow}
\usepackage{colortbl}
\usepackage{tabu}
\usepackage{pifont}

\usepackage[breaklinks=true,bookmarks=false]{hyperref}

\cvprfinalcopy % *** Uncomment this line for the final submission

\def\cvprPaperID{****} % *** Enter the CVPR Paper ID here

\begin{document}

%%%%%%%%% TITLE
\title{3rd Place Solution to ``Google Landmark Retrieval 2020"}

\author{Ke Mei$^\mathcal{y}$$^*$, Lei Li$ ^\mathcal{y}$\thanks{This work was done during the internship of K. Mei and L. Li in WeChat AI.}\\
Beijing University of Posts and Telecommunications\\
Beijing, China \\
{\tt\small \{raykoo, lei.li\}@bupt.edu.cn}
\and
Jinchang Xu$^\mathcal{y}$, Yanhua Cheng\thanks{The authors contributed equally.}, Yugeng Lin\\
WeChat AI\\
Beijing, China\\
{\tt\small \{maxwellxu, breezecheng, lincolnlin\}@tencent.com}
}

\maketitle
%\thispagestyle{empty}

%%%%%%%%% ABSTRACT
\begin{abstract}
   Image retrieval is a fundamental problem in computer vision. This paper presents our 3rd place detailed solution to the Google Landmark Retrieval 2020 challenge. We focus on the exploration of data cleaning and models with metric learning. We use a data cleaning strategy based on embedding clustering. Besides, we employ a data augmentation method called Corner-Cutmix, which improves the model's ability to recognize multi-scale and occluded landmark images. We show in detail the ablation experiments and results of our method.
\end{abstract}

%%%%%%%%% BODY TEXT
\section{Introduction}

Image retrieval is a fundamental computer vision task, which is to rank images according to their relevance to a query image. Unlike image classification or detection tasks like ImageNet \cite{deng2009imagenet}, COCO \cite{lin2014microsoft}, and other large-scale datasets, image retrieval is still evaluated on small datasets like Oxford5k \cite{philbin2007object} and Paris6k \cite{philbin2008lost}. Limited by the domain of small data, it is difficult to verify that existing methods are generalized on large-scale data sets and applied to real-world challenges. Google Landmarks Dataset v2 (GLDv2) is a new benchmark for large-scale, fine-grained image retrieval in the domain of human-made and natural landmarks \cite{weyand2020google}. 

This paper presents our 3rd place detailed solution to the Google Landmark Retrieval 2020 challenge\footnote{\href{https://www.kaggle.com/c/landmark-retrieval-2020}{https://www.kaggle.com/c/landmark-retrieval-2020}}. This year's challenge is structured in a representation learning format: rather than creating a submission file withe retrieved images, we will submit a model that extracts a feature embedding for the images. This rule means we can't use any post-processing methods such as DBA \cite{arandjelovic2012three} , QE \cite{radenovic2018fine} and re-rank. Therefore, we focus on the exploration of data preprocessing and model methods.

In our solution, we were inspired by the first place solution of the 2019 Challenge \cite{yokoo2020two} and used a data cleaning strategy based on embedding clustering. Besides, we came up with an unconventional data augmentation called Corner-Cutmix and technology based on metric learning. On this basis, we also tried a series of backbone networks.

\begin{figure}
\begin{center}
\includegraphics[width=3.2in]{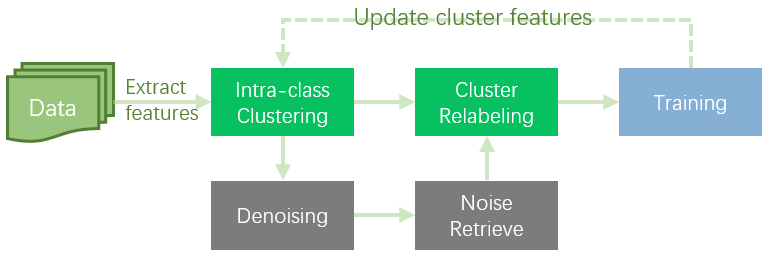}
\end{center}
  \caption{Pipeline of our data cleaning.}
\label{fig:data_cleaning}
\end{figure}

\section{Method}
This section focuses on our exploration of data and models. In section 2.1, we describe the details of how we use the model trained by GLDv2 clean \cite{weyand2020google} to re-clean the dataset by clustering images in GLDv2 and use Corner-Cutmix to do data augmentation. In section 2.2, we introduce the models and other tricks that we used in this challenge.

\subsection{Data Cleaning and Augmentation}

\noindent\textbf{Data Cleaning with Embedding Cluster.} Although GLDv2 clean has achieved excellent results in the challenges of the previous year\footnote{\href{https://www.kaggle.com/c/landmark-retrieval-2019}{https://www.kaggle.com/c/landmark-retrieval-2019}}, the reduction of data scale is still too serious. This reduces noise while reducing the amount of effective data and categories, which we believe can be avoided. 

\begin{figure}
\begin{center}
\includegraphics[width=3.2in]{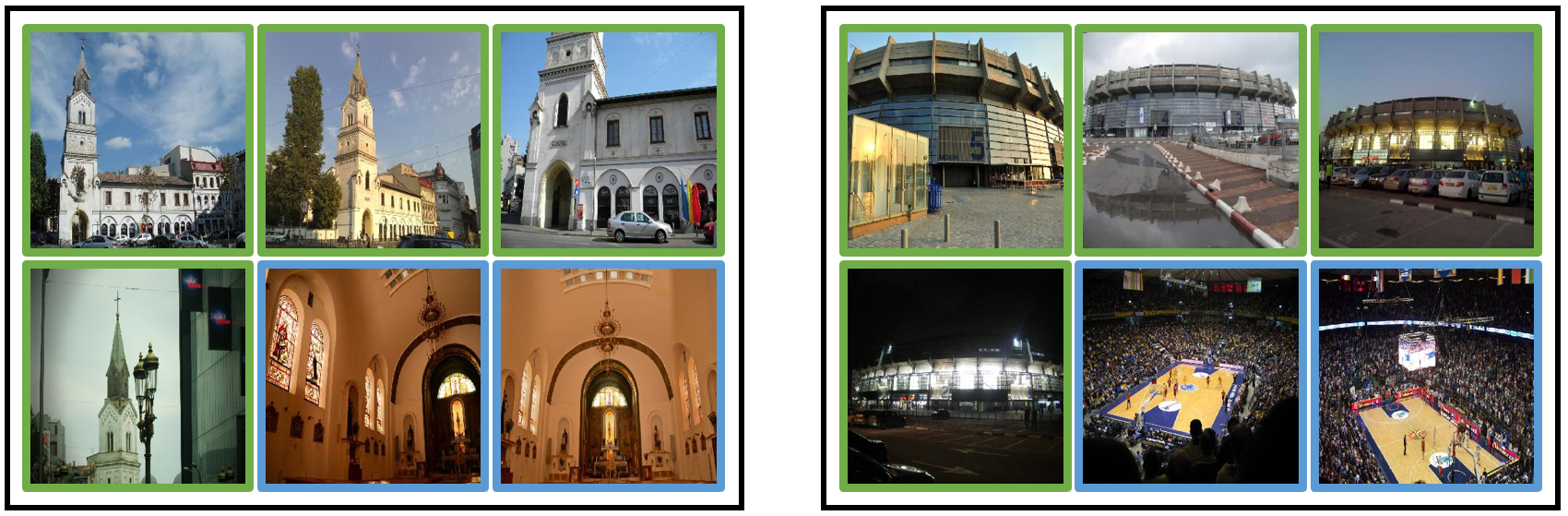}
\end{center}
  \caption{Visualization after data cleaning. The images in the black frame originally belong to the same category. These two sets of pictures are typical indoor (blue) outdoor (green) scenes. It is divided into two new categories by clustering algorithm to avoid the model being confused by it during training.}
\label{fig:data_visual}
\end{figure}

Like other competitors, we first use the GLDv2 clean version as the training set to train the model. Compared with the original GLDv2, this clean version has cleaner data and less noise to make the model easier to converge, which helps us quickly get an initial stable model. Based on this model, we extract the features of all GLDv2 and generate 512-dimensional embedding vectors for each images. As shown in Fig. \ref{fig:data_cleaning}, we use the density-based clustering algorithm DBSCAN \cite{ester1996density} to cluster embeddings of the same category. For each original category, after clustering, it can be automatically divided into several clusters, and noise data (data deviating from the cluster) can be detected. As shown in Fig. \ref{fig:data_visual}, the indoor and outdoor scenes in the original category can be easily divided into two clusters, which do not belong to the same category in terms of visual characteristics. We directly define each cluster as a new category. For the noise data, we used a more relaxed threshold to re-cluster separately, so as to retrieve a part of the noise as the long tail category to fully mine the available data.

After cleaning, we get a new version of the data set, called GLDv2 cluster, it has 2.8M images, and about 210k categories. Retraining on this new version of the dataset, we can obtain a better performing model. However, we have to remind that using this new model to update the clustering feature of the data, repeating the above cleaning process may not effectively improve the quality of the data set, and there may be a risk of overfitting.

~

\begin{figure}
\begin{center}
\includegraphics[width=3.2in]{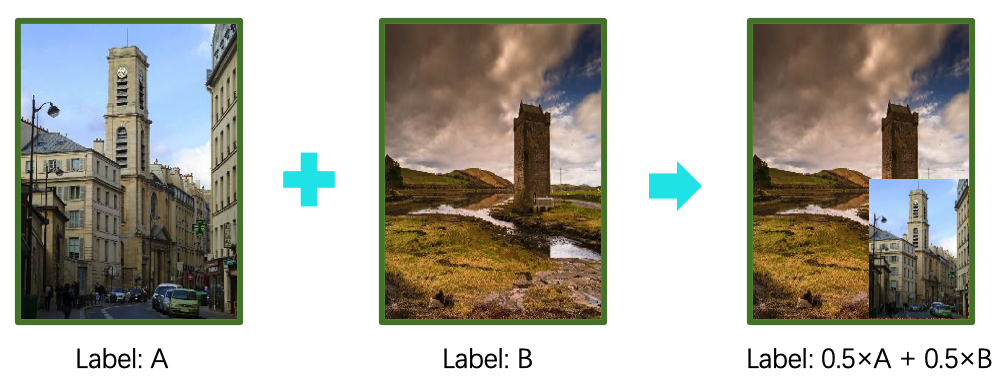}
\end{center}
  \caption{Illustrator of corner-cutmix method. For image A and image B, resize image A and cover the corners of image B to generate a new mixed image.}
\label{fig:data_corner-cutmix}
\end{figure}

\noindent\textbf{Data Augmentation with Corner-Cutmix.} In the landmark retrieval task, the same landmark exhibits multi-scales and occlusions due to different viewing angles and shooting distances. In addition to basic data augmentation strategies such as random resize and random crop, we propose a mixed data augmentation strategy called corner-cutmix, adapted from \cite{yun2019cutmix}. As shown in Fig. \ref{fig:data_corner-cutmix}, we mix two different images A and B by randomly overlay image A on a corner of image B to generate a new mixed image as training data, and simply assign it a new label with a weight of 0.5A and 0.5B. 

\begin{figure}
\begin{center}
\includegraphics[width=3.2in]{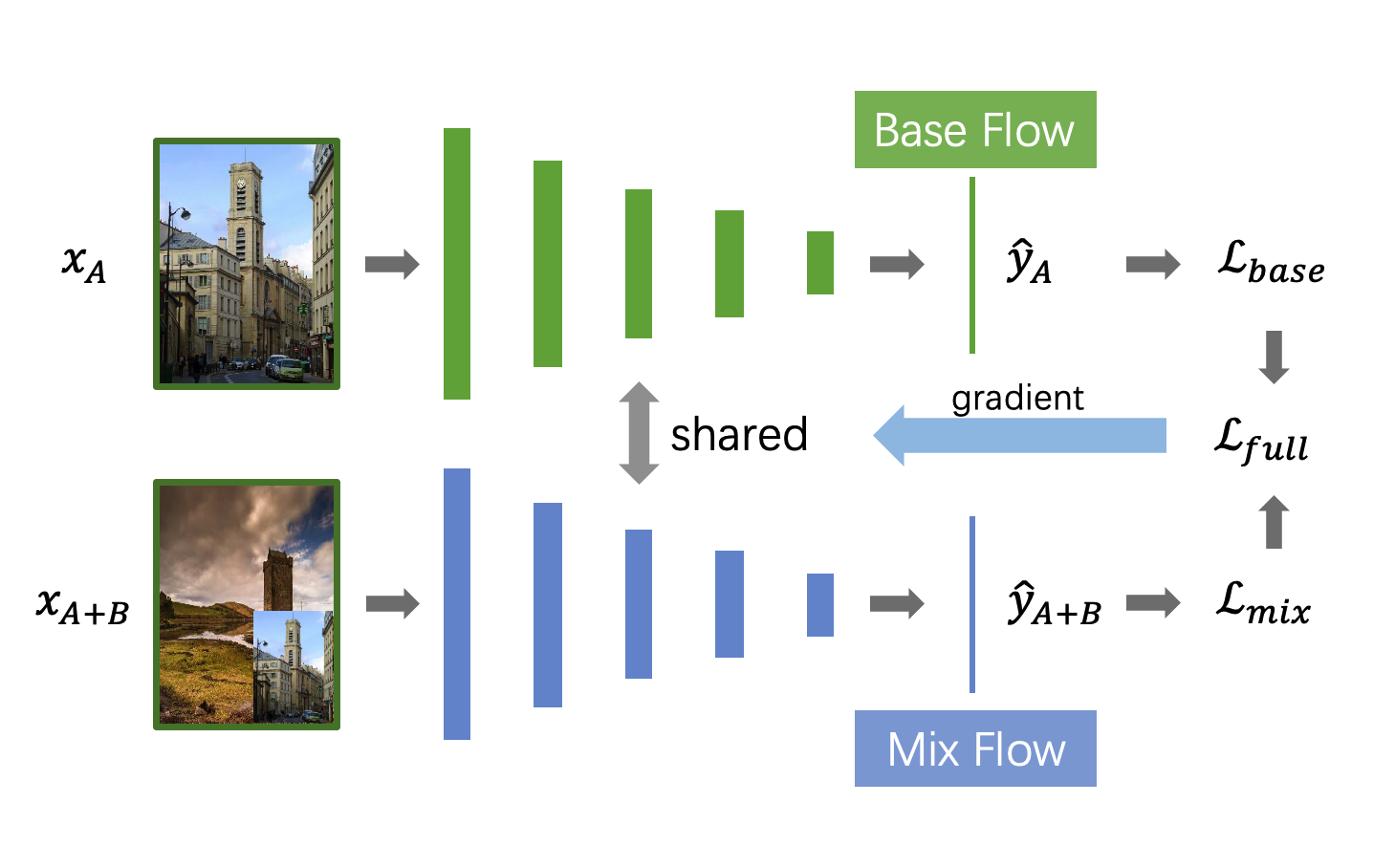}
\end{center}
  \caption{Network structure. The dual-stream network with shared weight. The basic image and the mixed image are input separately.}
\label{fig:data_network}
\end{figure}

Specifically, we use a dual-stream method to input the basic image $
x_{A} $ and mixed image $ x_{A+B} $, calculate the cross entropy loss of the two streams and return the gradient to update the network. In this way, while the network can benefit from the mixed images, it can also maintain the ability to analyze the original image and speed up the network convergence. As shown in Fig. \ref{fig:data_network}, the output of the two streams are $ \hat{y}_{A} $ and $ \hat{y}_{A+B} $, and the cross entropy loss of the two streams $ \mathcal{L}_{base} $ and $ \mathcal{L}_{mix} $. Among them, $ \mathcal{L}_{mix} $ is as follows:

$$
\mathcal{L}_{mix} = 0.5 \times CE(\hat{y}_{A+B}, y_A) + 0.5 \times CE(\hat{y}_{A+B}, y_B)
$$

\noindent Where $ CE\left ( \cdot \right ) $ represents cross entropy loss. For the mix flow, the labels of the two images are mixed. We simply fixed the weights of the two labels, instead of using the random value in \cite{yun2019cutmix}. Please refer to \cite{yun2019cutmix} for more detailed explanation.

\subsection{Modeling}
\noindent\textbf{Backbone Network.} As we all know, the backbone network is the most important factor that affects image feature representation. We tried various backbone networks with different architectures, such as ResNet152 \cite{he2016deep}, ResNest101 \cite{xie2017aggregated}, ResNest200 \cite{zhang2020resnest}, SE-ResNeXt101 \cite{hu2018senet} , Inception-v4 \cite{szegedy2016inception}, Efficientnet-b7 \cite{tan2019efficientnet}, etc. Limited by the short competition period, we have no way to complete the training to verify all the above models, so we finally choose ResNest200 and ResNet152 as our final submitted models. All our models use a two-stage training strategy: 1) first train on GLDv2 clean, 2) then finetune on the clustered GLDv2 cluster until the model converges. 

An interesting conclusion is that we found that some networks with very good performance on ImageNet have not satisfactory results in landmark retrieval tasks. Under the same time limit, a deeper network will have better performance than a wider network. For example, in our experiment, ResNet152, a simple backbone network, has better performance than ResNest101, SE-ResNeXt101, and Inception-v4 which have attention modules, group convolutions or inception modules, but ResNet152's performance is not as good as these networks on ImageNet. In addition, we did not use generalized mean-pooling (GeM) \cite{radenovic2018fine} but used the more general global average pooling (GAP) as the pooling method. GeM did not show obvious advantages in our experiments.

~

\noindent\textbf{Dimensionality Reduction.} After the GAP layer, we added a 1$\times$1 convolutional layer to reduce the dimension to 512, and then added batch Normalization, and finally got the output of the network after L2 normalization.

~

\noindent\textbf{Loss Function.} For this fine-grained retrieval task, we use cosine-softmax based loss, which is more able to separate the features of different categories than ordinary softmax loss. In this work, we first use ordinary softmax loss to pre-train a model on GLDv2 clean, and then use ArcFace \cite{deng2019arcface} with a margin of 0.3 and scale of 30 to finetune it.

~

\noindent\textbf{Model Ensemble.} In the final submission version, we ensemble the two best models (ResNet152 and ResNest200) by simply concatenate the output of the two models to form a 1024-dimensional vector as the final output.

\section{Experimental Result}
This section introduces our detailed experimental parameters and results of different models. We also add a part of the ablation experiment results of our method. Due to time and resource constraints, we cannot provide detailed ablation experiments.

\subsection{Training Settings}
Our implementation is based on the Pytorch 1.1 framework \cite{paszke2019pytorch}, using 8 NVIDIA Tesla P40 for training. The images are resized to 448$\times$448 before data augmentation. We use SGD \cite{bottou2010large} as the optimizer, the initial learning rate is set to 0.01, and lambda poly is used to adjust the learning rate. For most models, we first train 24 epochs on the GLDv2 clean dataset, and then train 12 epochs on the GLDv2 cluster dataset. For ResNest200, it is 12 epochs and 6 epochs. 

\subsection{Ablation Experiment}
We mainly verify the effectiveness of the dataset cleaned by clustering as well as the data augmentation with cornor-cutmix. Limited by the resource constraints, we only verify on different backbones, but we believe that this conclusion is commonly used with other backbones.

\begin{table}[htb]
\centering
\resizebox{0.9\linewidth}{!}{
\begin{tabular}{lcccrr}  
\toprule
Model & Loss & Data & Cutmix & Public & Private \\
\midrule
ResNest101 & S & clean & - & 0.305 & 0.266\\
ResNest101 & A & clean & - & 0.315 & 0.276\\
\midrule
ResNet152 & A & clean & - & 0.329 & 0.293 \\
ResNet152 & A & cluster & - & 0.356 & 0.321\\
\midrule
ResNest200 & A & cluster & - & 0.364 & 0.327\\
ResNest200 & A & cluster & \ding{51} & 0.374 & 0.334\\
\bottomrule\\
\end{tabular}
}
\label{table:ablation}
\caption{Results of ablation study. S means softmax loss, and A means using ArcFace. mAP@100 is used for evaluation.}
\end{table}

As shown in Table 1, we demonstrate the effectiveness of ArcFace, GLDv2 cluster and Corner-Cutmix. Under the same backbone, they can steadily improve performance.

\subsection{Comparison of Backbone Networks}
Due to time constraints, we have not completed training on some backbones (such as Inception v4, EfficientNet-b7, etc.). In the case of controlling other variables to be consistent, we compare the performance of several backbone networks as shown in Table 2.

\begin{table}[htb]
\centering
\resizebox{0.9\linewidth}{!}{
\begin{tabular}{llcrrr}  
\toprule
\# & Backbone & Test size & Public & Private & Time \\
\midrule
1 & ResNet152 & 448$\times$448 & 0.356 & 0.321 & 26.3\\
2 & SE-ResNeXt101 & 448$\times$448 & 0.344 & 0.307 & 59.8\\
3 & ResNest101 & 448$\times$448 & 0.328 & 0.294 & 41.2\\
4 & ResNest200 & 448$\times$448 & 0.364 & 0.327 & 74.1\\
\midrule
5 & Emsamble 1+4  & 448$\times$448 & 0.376 & 0.342 & -\\
6 & Emsamble 1+4 & 512$\times$512 & 0.381 & 0.346 & - \\
\bottomrule\\
\end{tabular}
\label{table:backbone}
}
\caption{Comparison of different backbone networks, among which \#6 is the model we finally submitted. ``Time" means the mean test time (ms) of 200 images.}
\end{table}

\section{Conclusion}
In this paper, we describe our winning solution in the Google Landmark Retrieval 2020 challenge. We use a data cleaning strategy based on embedding clustering. Besides, we employ a data augmentation method called corner-cutmix, which improves the model's ability to recognize multi-scale and occluded landmark images.

{\small

\bibliographystyle{ieee_fullname}
\bibliography{egbib}
}

\end{document}